\documentclass{article}
\usepackage{arxiv}
\usepackage[utf8]{inputenc} 
\usepackage[T1]{fontenc}    
\usepackage{hyperref}       
\usepackage{url}            
\usepackage{booktabs}       
\usepackage{amsfonts}       
\usepackage{nicefrac}       
\usepackage{microtype}      
\usepackage{lipsum}		
\usepackage{graphicx}
\usepackage[square,sort,comma,numbers]{natbib}
\usepackage{doi}
\usepackage{times}
\usepackage{epsfig}
\usepackage{graphicx}
\usepackage{amsmath}
\usepackage{amssymb}
\usepackage{multirow}
\usepackage{algorithm}
\usepackage{subfig}
\usepackage{microtype}
\usepackage{stfloats}
\usepackage{float}
\usepackage{mathrsfs}
\usepackage{makecell}
\usepackage{colortbl}
\usepackage[noend]{algpseudocode}
\usepackage{pifont}
\newcommand{\cmark}{\ding{51}}%
\newcommand{\xmark}{\ding{55}}%
\usepackage{soul}
\usepackage{enumitem}
\usepackage[
  separate-uncertainty = true,
  multi-part-units = repeat
]{siunitx}
\newcommand{\etal}{\textit{et al.}}

\title{Sparse Adversarial Video Attacks with Spatial Transformations
}

\author{
  Mu, Ronghui \\
  Computing and Communications, \\
  Lancaster University \\
  Lancaster\\
  \texttt{ronghui.mu@lancaster.ac.uk} \\
   \And
  Ruan, Wenjie \\
  Computer Science, \\
  University of Exeter, \\
  Exeter, UK\\
  \texttt{w.ruan@exeter.ac.uk} \\
   \AND
   Soriano Marcolino,Leandro \\
  Computing and Communications, \\
  Lancaster University \\
  Lancaster\\
 \texttt{l.marcolino@lancaster.ac.uk} \\
  \And
  Ni,Qiang \\
  Computing and Communications, \\
  Lancaster University \\
  Lancaster\\
  \texttt{q.ni@lancaster.ac.uk} \\
}
\begin{document}
\maketitle

\begin{abstract}

In recent years, a significant amount of research efforts concentrated on adversarial attacks on images, while adversarial video attacks have seldom been explored. We propose an adversarial attack strategy on videos, called DeepSAVA.  Our model includes both additive perturbation and spatial transformation by a unified optimisation framework, where the structural similarity index measure is adopted to measure the adversarial distance. We design an effective and novel optimisation scheme which alternatively utilizes Bayesian optimisation to identify the most influential frame in a video and Stochastic gradient descent (SGD) based optimisation to produce both additive and spatial-transformed perturbations. Doing so enables DeepSAVA to perform a very sparse attack on videos for maintaining human imperceptibility while still achieving state-of-the-art performance in terms of both attack success rate and adversarial transferability. Our intensive experiments on various types of deep neural networks and video datasets confirm the superiority of DeepSAVA.

\end{abstract}


\section{Introduction}

In the past decade, Deep Neural Networks (DNNs) have demonstrated their outstanding performance in various domains, such as image classification \cite{medical,wu2018game}, text analysis \cite{9074228}, speech recognition \cite{fohr:hal-01484447}, and object detection \cite{fohr:hal-01484447,zhang2021fooling}. Despite their huge success in these tasks, recently some researchers have shown that DNNs are surprisingly vulnerable to adversarial attacks \cite{szegedy2014intriguing,carlini2017evaluating,yuan2018adversarial,RHK2018}, e.g., adding a small human-imperceptible perturbation to an input image can fool DNNs, enabling the model to make an arbitrarily wrong prediction with high confidence \cite{szegedy2014intriguing,RWSHKK2019}. This is raising serious concerns about the readiness of deep learning models, especially on safety-critical applications such as face authentication \cite{mohammadi2017deeply}, surveillance systems \cite{pan2018deep}, and medical applications \cite{medical}. Hence, it is of vital importance to investigate the performance of DNNs on adversarial examples and evaluate their robustness in an adversary environment~\cite{xu2020towards,sun2018concolic,ruan2021adversarial}.

As such, significant research efforts have emerged to assess the robustness of DNNs under adversarial attacks, notable works include Fast Gradient Sign Methods (FGSM) \cite{yuan2018adversarial}, C\&W attack \cite{carlini2017evaluating}, DeepFool \cite{moosavidezfooli2016deepfool}, and JMSA \cite{wiyatno2018maximal}. 
These attack strategies primarily concentrate on image-related tasks, yet the adversarial robustness of deep learning models on videos has not been comprehensively explored. Recently, a number of works \cite{wei2019sparse,naeh2020flickering,jiang2019blackbox,yan2020sparse} are aware of the values of the adversarial attacks on videos. Theoretically, attacking videos is more challenging compared with images because videos contain temporal information. So video attack not only requires to achieve minimal adversarial distance but also needs to perturb as few frames as possible. 
As such, identifying the most {\em effective} frame(s) and generating {\em competitive} perturbation upon those frame(s) are of huge importance to the success rate of the attack. Another important consideration is the efficiency. As perturbing each frame of the video is time-consuming, we expect to perform the influential-frame identification and adversarial perturbation simultaneously so we can maintain human imperceptibility and achieve high attacking success rate. In practice, DNNs processing videos are widely applied in real systems such as video surveillance \cite{pan2018deep}, action recognition \cite{kalfaoglu2020late} and autonomous driving~\cite{wu2021adversarial}. In particular, most of those applications directly relate to the decisions concerning property security or human health and safety.
As a result, investigating adversarial samples on videos is urgently needed. However, to achieve a high-performance adversarial attack on videos, the following challenges need a careful treatment:
\begin{figure}\small
\centering
\subfloat[{\em SSIM=1 $l_{1,2}$=0}]{
\begin{minipage}{2cm}
\centering
\includegraphics[width=2cm]{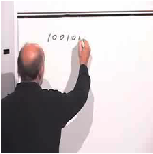}
\end{minipage}%
}%
\subfloat[{\em SSIM=0.92 $l_{1,2}=3.72$ }]{
\begin{minipage}{2.5cm}
\centering
\includegraphics[width=2cm]{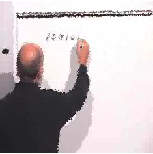}
\end{minipage}%
}%
\subfloat[{\em SSIM=0.92 $l_{1,2}$=4.12}]{
\begin{minipage}{2.5cm}
\centering
\includegraphics[width=2cm]{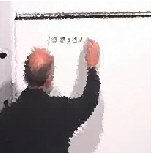}
\end{minipage}%
}%
\subfloat[{\em SSIM=0.91 $l_{1,2}$=4.01 }]{
\begin{minipage}{2.5cm}
\centering
\includegraphics[width=2cm]{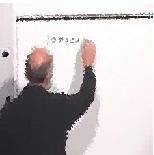}
\end{minipage}%
}%
\subfloat[{\em SSIM=0.92 $l_{1,2}$=3.91 }]{
\begin{minipage}{2.5cm}
\centering
\includegraphics[width=2cm]{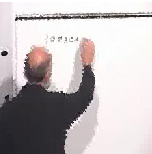}
\end{minipage}%
}%
\caption{SSIM and $l_{1,2}$ norm distance for: (a) original image; Perturbed images: (b) noise; (c) scaling + noise; (d) rotation + noise; (e) rotation+scaling + noise.}
\label{Fig:SSIMandlp}
 
\end{figure}

{\em Maintaining the fidelity of the produced adversarial examples with real-world scenarios:} One important criterion for adversarial attacks is that the perturbed example should resemble a real-world instance as close as possible. Current video attack strategies all adopt the $l_p$-norm metric to measure the fidelity of the perturbed examples. Although $l_p$-norm is effective to capture noise contamination, it is sensitive to naturally-occurred transformations such as rotation, spatial shift, and scaling \cite{4775883}. Taking Figure \ref{Fig:SSIMandlp} as an example, a slight rotation or scaling of pixels will lead to an obvious difference in $l_p$-norm distance. Thus, attacks constrained by $l_p$-norm cannot capture some spatial transformations that naturally happen in a real-world scenario such as the shaking, vibration, or rotation of a camera.

{\em Achieving a high attacking success rate without compromising the human imperceptibility:} Different to static images, videos contain sequential data structure and change dynamically with the temporal dimension. Hence, the existing attack strategies designed for images are not directly applicable to videos. Perturbing all frames of a video, although could achieve a high fooling rate, is time-consuming and also potentially compromises human imperceptibility. Thus, perturbing as few frames as possible while maintaining a high attacking success rate is highly desired on adversarial video attacks, which can be tackled by sparse attacks.

{\em Enabling the adversarial video attack to be effective across diverse types of DNNs:} In a real-world scenario, we may not be able to access the parameters, structures, or even datasets of a pre-trained DNN. Thus, similar to adversarial attacks on images, a strong adversarial transferability that can work across diverse unseen models is desirable. However, unlike DNNs for images that are without temporal structure, video models are more complicated and include diverse neural units for recurrent operations. Hence, achieving a satisfying adversarial transferability across unseen models is also a non-trivial challenge.
    
As a result, we provide a pioneering exploration to deal with the above challenges. We propose an adversarial video attack for DNNs, called DeepSAVA, which can {\em i)} capture a wide range of adversarial instances including both noise contamination and various spatial transformations; {\em ii)} achieve sparse attack, i.e., only perturbing very few frames of a video while still achieving a state-of-the-art attack success rate; and {\em iii)} obtain better adversarial transferability across various recurrent models compared with baseline methods. 

In summary, there are three key {\bf contributions} in DeepSAVA:

{\bf Structural Similarity Index Measure (SSIM):} instead of $l_p$-norm, we adopt the SSIM metric in the loss function to constrain the distance between adversarial and clean videos. As demonstrated by the community of Image Quality Assessment, SSIM is an alternative signal fidelity measure that is superior to $l_p$-norm on some applications where human perceptual criterion matters \cite{4775883}. As Figure \ref{Fig:SSIMandlp} shows, SSIM is less sensitive to both noise and spatial transformations such as rotation and scaling, which is more resemble human perception.
    
{\bf Combination of additive perturbation and spatial transformation:} we are the first work to combine additive and spatial-transformed perturbation for video attacks. According to the image attack used spatial transformation perturbation \cite{Xiao2018SpatiallyTA,zhang2020generalizing}, changing the positions of pixels could improve perceptual realism and make it locally smooth. In DeepSAVA, we introduce a new term in the loss function for optimising both additive and spatial transformation perturbation. With a proper SSIM-based constraint, we could produce strong perturbations combined with additive and spatial transformation. Such combined perturbation enables DeepSAVA to achieve successful attacks by just perturbing one frame. 
    
{\bf Novel alternating optimisation strategy:}  we are also the first work that uses Bayesian optimisation (BO) to choose the most critical frames of the video in attacks. To achieve a video attack that can perturb as few frames as possible, we design an alternating optimisation strategy that can effectively identify the key frames via BO and then initiate additive and spatial-transformed perturbations on the selected key frames by stochastic gradient descent (SGD) based optimiser. Such an alternating process happens in each iteration of the optimisation until key frames are found. Combining the above two ingredients, the proposed novel optimisation strategy could achieve a better fooling rate than baselines.

\begin{figure*}
\begin{center}
\includegraphics[width=10cm,height=5.5cm,
]{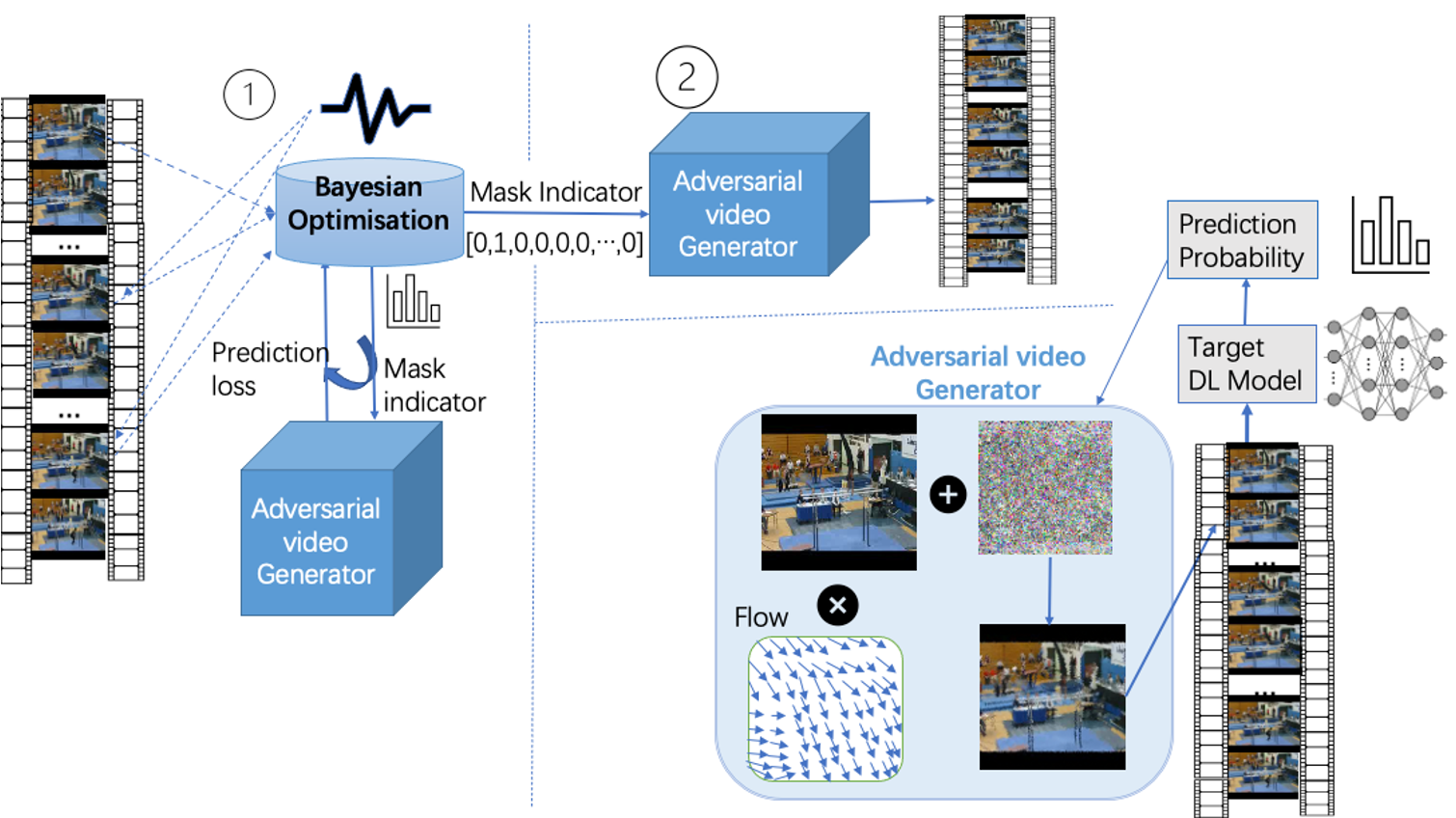}
\end{center}
\vspace{-4mm}
  \caption{Overview of DeepSAVA: the key frame is alternatively identified by BO; the additive and spatial-transformed perturbations are applied to the selected frame to generated adversarial examples.}
\label{fig:flow}
\end{figure*}

The flow chart of our method is illustrated in Figure \ref{fig:flow}. We anonymously release the code of DeepSAVA \footnote[1]{https://github.com/TrustAI/DeepSAVA} and generated adversarial videos across multiple models \footnote[2]{https://www.youtube.com/channel/UCBDswZC2QhBhTOMUFNLchCg}.

\section{Related Work}

 \textbf{Video Action Recognition Models:}  The video classification task primarily focused on action recognition \cite{kong2018human}. The works on video classification using DNNs are developed in two ways: using 2D or 3D-based convolution neural networks (CNN). Since the CNNs have obtained state-of-the-art performance on image classification, Karpathy \etal \cite{6909619} first proposed to use 2D CNN to classify each frame of the video. Szegedy \etal then developed the Inception-v3 \cite{szegedy2014going,szegedy2015rethinking}, which is commonly used as a baseline classification model. As 2D-CNNs use incomplete video information, some works added layers containing temporal information such as LSTM to integer CNN features extracted over time which is referred to as CNN+LSTM model \cite{Nguyen_2015_CVPR,donahue2016longterm}. As for the 3D CNNs \cite{tran2015learning}, it can learn temporal features from videos by inputting all frames in three dimensions directly. Vadis \etal \cite{carreira2018quo} proposed a two-stream inflated 3D CNN (I3D) to build the 2D kernel first and then merge the pooling layer and kernel into a 3D network. By pre-training the I3D on Kinetics Dataset, it could reach state-of-the-art performance on recognising UCF101 and HMDB51 action video datasets. 

 \textbf{Adversarial attack on images:} The adversarial attack on images has been explored extensively recently. Szegedy \etal \cite{szegedy2014intriguing} first proposed to add visually imperceptible noise on the images to mislead pre-trained CNN to give the wrong prediction label. Goodfellow \etal \cite{goodfellow2014explaining} proposed to use a gradient-based approach, the fast gradient sign method (FGSM), to generate adversarial examples. DeepFool \cite{moosavidezfooli2016deepfool} is then proposed to find the minimal perturbation by iteratively linearizing the loss function. Other gradient-based optimisation algorithms to generate perturbation were also proposed \cite{carlini2017evaluating,liu2017delving,tanay2016boundary,xiao2018generating}. These works mentioned above only apply additive perturbation on pixels. 
 Some works {\cite{Xiao2018SpatiallyTA,wong2020wasserstein,laidlaw2019functional,laidlaw2021perceptual,jordan2019quantifying}} use a functional perturbation which is non-additive-only perturbation like spatial transformation. These perturbation slightly perturb the location of pixels. Some works such as \cite{jordan2019quantifying,zhao2020large,gragnaniello2020perceptual} also utilize other types of metrics such as SSIM to quantify the human perception, but none of them explored the SSIM-guided spatial transformation. For more details on adversarial attacks, please refer to our recent survey~\cite{huang2020survey} and tutorial~\cite{ruan2021adversarial,berthier2021tutorials}.
 
 \begin{table*}\footnotesize
\begin{center}
\scalebox{0.9}{
{
\begin{tabular}{|c|c|c|c|c|c|c|c|c|} 
\hline
{} & \makecell[c]{Flickering\\ \cite{naeh2020flickering}} & \makecell[c]{RL\\ \cite{yan2020sparse}}& \makecell[c]{Heuristic\\ \cite{wei2020heuristic}} & \makecell[c]{Append\\  \cite{chen2019appending}} & \makecell[c]{BlackBox \\ attack \cite{jiang2019blackbox}}& \makecell[c]{GAN-based\\attack \cite{Li_2019}} & \makecell[c]{Sparse \\ Attack \cite{wei2019sparse}} & \makecell[c]{Deep \\ SAVA} \\
\hline\hline
Similarity metric  & $l_p$ & $l_{p}$ & $l_{1}$ & $l_{\infty}$ & $l_{\infty}$ & $l_{p}$& $l_{2,1}$ & SSIM \\
\hline
Spatial-transformed perturbation & \xmark & \xmark & \xmark & \xmark & \xmark & \xmark & \xmark & \cmark\\
\hline
Additive Perturbation & \cmark & \cmark & \cmark & \cmark & \cmark & \cmark & \cmark & \cmark\\
\hline
Identify Key Frames & \xmark & \cmark & \cmark & \xmark &\xmark &\xmark &\xmark & \cmark\\
\hline
Transferability Study & \xmark & \xmark & \xmark & \cmark & \cmark &\xmark&\cmark & \cmark \\
\hline
Sparse Attack & \xmark & \cmark & \cmark & \xmark & \xmark& \xmark& \cmark & \cmark \\
\hline

\end{tabular}}}
\end{center}

\caption{Comparison with related work in different aspects.}
\label{tabel:related}

\end{table*}

 \textbf{Adversarial attack on videos:} Wei \etal \cite{wei2019sparse} claimed that they are the first to attack videos. Instead of attacking each frame of a video, they apply additive perturbations on randomly selected frames and use $l_{2,1}$ norm to guide the gradient-based optimisation and evaluated the performance on the CNN+LSTM model. Li \etal \cite{Li_2019} used a GAN network to generate offline universal perturbations for each frame. Chen \etal \cite{chen2019appending} proposed to append a noise frame to the end of videos, which is obtained based on all videos. Naeh \etal \cite{naeh2020flickering} applied flickering temporal perturbations on each frame to generate universal perturbations for the I3D model. Jiang \etal \cite{jiang2019blackbox} were the first to propose a black-box approach to attack videos. Wei \etal \cite{wei2020heuristic} proposed to use a heuristic method and Yan \etal \cite{yan2020sparse} used a reinforcement learning algorithm to select the key frames to perform black-box attack. However, these works only applied additive perturbation based on $l_p$-norm distance. In Table \ref{tabel:related}, we compare our method with existing related works on video attacks in six aspects. Our work applies the SSIM-guided non-additive perturbation on selected frames to generate adversarial videos efficiently. We also propose a novel alternating optimisation strategy to select the key frames.

\section{Methodology}
\textbf{Problem Definition:} The video classifier is defined as \small{$J(\cdot ; \boldsymbol{\theta})$} with pretrained weights \small{$\boldsymbol{\theta}$}. The input clean video is defined as \small{$\mathbf{X} = (x_{1},x_{2},...,x_{T}) \in \mathbb{R}^{T \times W \times H \times C}$}, where \small{$T$} is the length of the video (number of frames), and \small{$W, H, C$} represents the width, height, and the number of channels of each frame; its adversarial video generated is represented as \small{$\hat{\mathbf{X}}$}. In order to obtain the adversarial example, the original video is perturbed by a spatial transformer \small{$\mathcal{S}$}, and additive noise $\mathcal{D}$. Given that the ground truth label of input video $\mathbf{X}$ is $y$, the objective function is:
\begin{equation}\small
  \arg \min \lambda\ell_{similar}(\hat{\mathbf{X}},{\mathbf{X}})-\ell_{adv}\left(\mathbf{1}_y, J(\hat{\mathbf{X}} ; \boldsymbol{\theta})\right),
\end{equation}
where $\mathbf{1}_y$ is the one-hot encoding of $y$; $\ell_{similar}$ is the similarity loss function to measure the distance between generated adversarial and original video; $\ell_{adv}$ is the loss function to measure the difference between ground truth and prediction label. The parameter $\lambda$ is set to balance these two loss terms.
Additionally, the cross-entropy is used to calculate the $\ell_{adv}$, which is proved to be effective in \cite{wei2019sparse}. 
\vspace{-3mm}
\subsection{Sparse Spatial Transform Adversarial Attack}
\begin{figure}
\centering
\subfloat[]%
{
\begin{minipage}{7cm}
\centering
\includegraphics[width=1\linewidth]{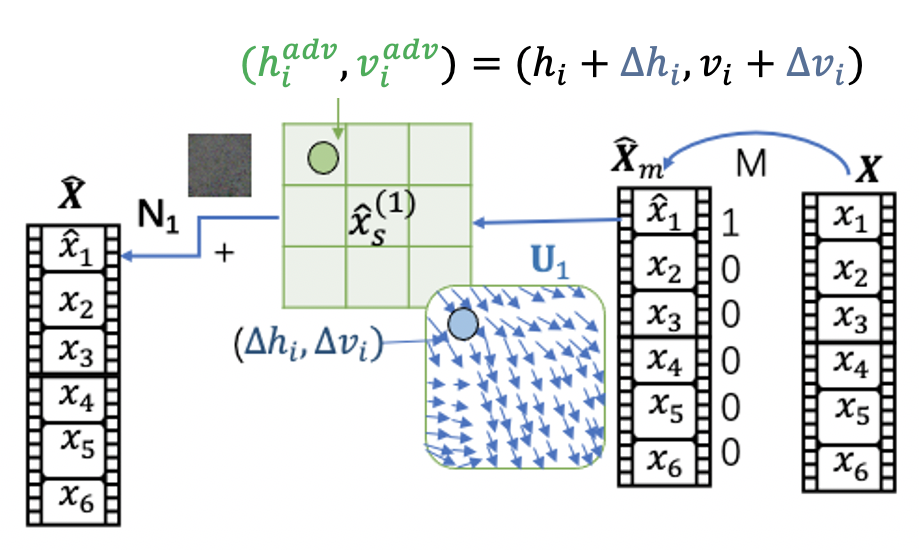}
\end{minipage}%
}%
\subfloat[]%
{
\begin{minipage}{7cm}
\centering
\includegraphics[width=1\linewidth]{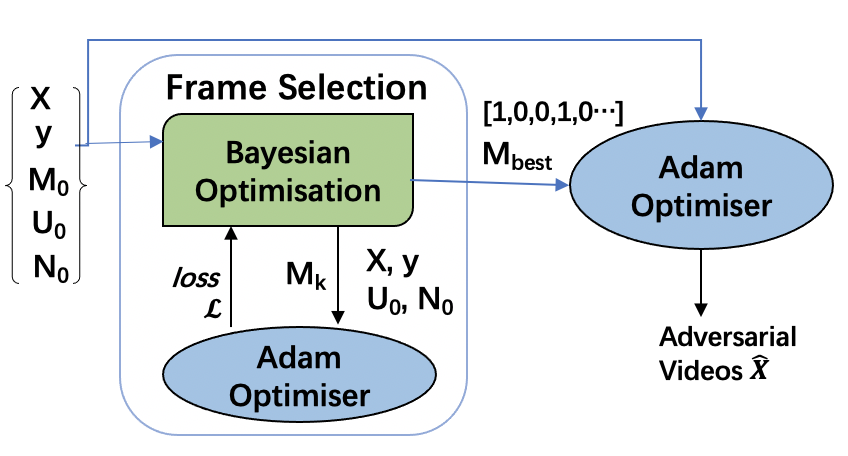}
\end{minipage}%
}%
\caption{(a) The process to perturb one frame of a clean video $\mathbf{X}$, where the first frame is masked to be perturbed by spatial flow vector $\mathbf{U}$ and noise $\mathbf{N}$. (b) The systematic optimisation process by using Bayesian Optimisation and Adam Optimiser.}

\label{fig:perturb}
\end{figure}

\textbf{Structural Similarity Index Measure (SSIM)}:
The SSIM was first proposed in \cite{wang2002universal}, and is detailed in \cite{wang2004image}. It measures the local similarities between the local pixels on three aspects: structures, contrasts, and brightness. As we mentioned before, the SSIM is less sensitive to the combined perturbation and more similar to human perception than $l_p$-norms \cite{4775883}. As the SSIM is differentiable with respect to the input variable (the definition and derivation process of SSIM are shown in Appendix A), we apply SSIM to calculate the similarity loss to constrain the perturbation during the optimisation process. The overall SSIM score for the video is calculated by summing up the SSIM loss over all frames.

\textbf{Sparse Attack:} The mask indicator \small{$M = (m_1, m_2, .., m_T) \in \mathbb{R}^T$} is used to choose the key frames in the video, where \small{$m_t \in\{\mathbf{0}, \mathbf{1}\}$} indicates whether the $t$-th frame is masked to be perturbed. The masked video \small{$\mathbf{X}_m$} is formed through the map function \small{$\mathcal{M}(M,\mathbf{X})$}, and then fed into the spatial transformer $\mathcal{S}$.

\textbf{Spatial Transformed Perturbation:} Given the $t$-th frame $x^t \in \mathbb{R}^{W \times H \times C}$ of input video $\mathbf{X}$, $x_{n}^t$ denotes the $n$-th pixel of $x^t$ and its location in the frame can be represented by a 2D coordinate \small{$(h_{n}^t,v_{n}^t)$}. The spatial transformer \cite{jaderberg2015spatial} $\mathcal{S}$ is a differentiable model composed by flow displacement vectors \small{$\mathbf{U} =((\Delta{\mathbf{H}^1},\Delta{\mathbf{V}^1}),(\Delta{\mathbf{H}^2},\Delta{\mathbf{V}^2}),...,(\Delta{\mathbf{H}^T},\Delta{\mathbf{V}^T})) \in \mathbb{R}^{T\times 2\times H \times W }$} (where \small{$\mathbf{H}^t=(h_{0}^t,h_{1}^t,...,h_{n}^t)$}, \small{$\mathbf{V}^t=(v_{0}^t,v_{1}^t,...,v_{n}^t)\in \mathbb{R}^{H \times W}$}), which is used to synthesize the 2D coordinate of adversarial videos. Suppose $\hat{x}_{n}^t$ with location $(\hat{h}_{n}^t,\hat{v}_{n}^t)$ is the adversarial example transformed from $x_{n}^t$, given its corresponding spatial displacement vector $(\Delta{h_{n}^t},\Delta{v_{n}^t})$, the new location of original pixel $x_{n}^{t}$ can be represented as $(h_{n}^t,v_{n}^t) = (\hat{h}_{n}^t+\Delta{h_{n}^t},\hat{v}_{n}^t+\Delta{v_{n}^t})$. 
Considering the sparse attack mask indicator $M$, we can represent the transformed adversarial video as \small{$\hat{\mathbf{X}}_S = \mathcal{S} (\mathbf{U}, \mathbf{X},M)$}. 

\textbf{Additive Perturbation:} The additive perturbation is the most common way to generate adversarial examples \cite{carlini2017evaluating,goodfellow2014explaining}. We define the additive model as $\mathcal{D}$ with parameter $\mathbf{N} \in \mathbb{R}^{T \times W \times H \times C}$. We combine spatial transformation and additive perturbation to generate adversarial videos as (illustrated in Figure \ref{fig:perturb} (a)): 
\small{$\hat{\mathbf{X}} = \mathcal{D}(\mathbf{N},\hat{\mathbf{X}}_S,M)= \mathbf{N}\cdot M + \hat{\mathbf{X}}_S$}.


\subsection{Novel Alternating Optimisation Strategy}
In this paper, we utilize the Bayesian Optimisation (BO) to select the most critical frames. As the frame selection is a discrete variable optimisation problem, we also tried other discrete optimisation techniques such as simulated annealing (SA) \cite{annealing} and genetic algorithms (GA) \cite{Whitley94agenetic}, but both spent about 200s to find the final result which is much longer than about 16s taken by BO. 

The generated adversarial video is formed as \small{$\hat{\mathbf{X}} = \mathbf{N}\cdot M + \mathcal{S}(\mathbf{U},\mathbf{X},M)$}. In this paper, the similarity loss $\ell_{similar}$ and adversarial loss \small{$\ell_{adv}$} in problem (1) can be expressed as \small{$\ell_{similar}(\hat{\mathbf{X}},{\mathbf{X}})= 1 - SSIM(\hat{\mathbf{X}},{\mathbf{X}}) = \mathcal{L}_s(\mathbf{N}, \mathbf{U}, \mathbf{X}, M)$} and \small{$\ell_{adv}\left(\mathbf{1}_y, J(\hat{\mathbf{X}} ; \boldsymbol{\theta})\right) = \mathcal{L}_a(\mathbf{N}, \mathbf{U}, \mathbf{X}, M)$}. Therefore, problem (1) can be simplified as: \small{$
\arg \min_{M,\mathbf{N},\mathbf{U}}\lambda\mathcal{L}_s(\mathbf{N}, \mathbf{U}, \mathbf{X}, M)- \mathcal{L}_a(\mathbf{N}, \mathbf{U}, \mathbf{X}, M)
$}.

As $M$ is a discrete binary vector, which makes problem (4) non-differentiable, we solve it systematically by a novel alternating optimisation strategy: at each iteration, we optimise $M$ by BO first; and then by fixing $M$, the problem becomes differentiable, which can be solved by Stochastic Gradient Descent (SGD) based optimisation. We choose the Adam optimiser \cite{kingma2014adam} because of its robust and fast convergence performance. This process repeats for a fixed number of iterations, continuously improving the solution via both techniques alternatively.

BO proposes sampling points from the search space through acquisition functions to obtain the reward of previous points. Expected improvement (EI) is applied as acquisition function $F$, which is widely employed in BO: 
\small{ $\mathbb{E} [\max \left(\mathcal{L}(M)-\mathcal{L}\left(M^{+}\right), 0\right)]$}
, where $\mathcal{L}(M)$ is the loss from Adam by fixing $M$; $\mathcal{L}\left(M^{+}\right)$ is the best value obtained so far and $M^+$ is its location. 

During the BO process, we will find the best mask indicator through several iterations. In the $k$-th iteration of BO, we first sample a candidate $M^k$ according to the acquisition function $F$.

Then, the corresponding loss $\mathcal{L}_k$ will be computed by the Adam, which will then affect the next sampled point $M^{k+1}$ for the next iteration.
When the BO reaches the maximum exploration number, the best $M$ with minimum loss will be fed into the Adam optimiser to generate the final adversarial video. The process is illustrated in Figure \ref{fig:perturb}(b).

 \begin{figure}
\centering
\begin{minipage}{8cm}
\centering
\includegraphics[width=8cm]{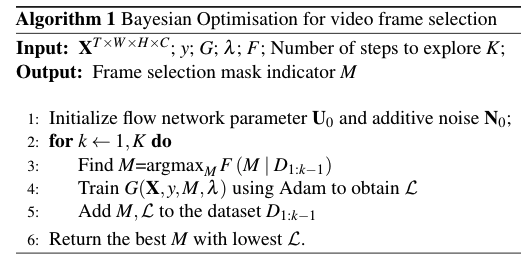}
\end{minipage}%
\begin{minipage}{8cm}
\centering
\includegraphics[width=8cm]{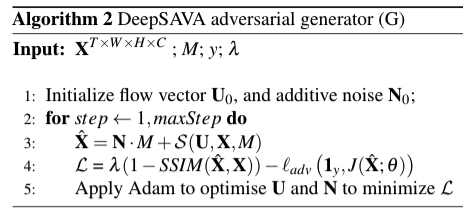}
\end{minipage}%
\label{Fig:frame}
\end{figure}

Algorithm 1 and 2 detail the BO selection and adversarial videos generation algorithms respectively. In Algorithm 1, the next sampling point $M$ is obtained by maximizing the acquisition function $F$ based on previous sampling data set ${D}_{1: k-1}$ (Line 3). After adversarial Generator (G) is optimized, the loss $\mathcal{L}$ for $M$ is calculated. Then the $M$ with its corresponding $\mathcal{L}$ are appended to the sampling pool $D$ to propose the next sampling point. In Algorithm 2, according to the optimised mask indicator $M$, the final flow vector $\mathbf{U}$ and additive noise $\mathbf{N}$ are optimised via Adam.

\begin{table}\footnotesize
\begin{center}
\begin{tabular}{|c|c|c|c|c|c|c|c|}
\hline
 & \multicolumn{3}{c|}{UCF101} & \multicolumn{3}{c|}{HMDB51} \\
\hline
Models&CNN+LSTM &I3D & Inception-v3&CNN+LSTM &I3D & Inception-v3 \\
\hline
Accuracy &74\%& 94.9\% & 71.2\%&43\%&80\%&47\%\\
\hline
\end{tabular}
\end{center}
\caption{Training accuracy of the classifiers to be attacked.}
\label{label:acc}

\end{table}

\section{Experiments}

\textbf{Dataset:} As action recognition video datasets are widely used in adversarial video attack studies, we choose two popular benchmark action recognition datasets to evaluate the performance of our method: UCF101 \cite{soomro2012ucf101} and HMDB51 \cite{kuehne2011hmdb}. Both datasets are realistic action recognition datasets. The UCF101 contains 13,320 videos with 101 categories such as playing instruments, body movements, human-object interaction. Similarly, the HMDB51 has around 7,000 videos within 51 categories related to body-motion and facial actions.

\textbf{Action Recognition Models:} We evaluate DeepSAVA on three classifiers: {\em Inception-v3}, a 2D-CNN based model \cite{szegedy2015rethinking}, which is widely used in the image recognition task with high accuracy; {\em I3D}, a 3D-CNN based model, pre-trained on Kinetics \cite{carreira2018quo}; {\em CNN with LSTM}, which is pre-trained on ImageNet to extract features from videos and then input these features to train the LSTM network. The training accuracy of all classifiers is shown in Table \ref{label:acc}.

\textbf{Baseline methods:} 
Two baseline methods are used for comparison, the Sparse \cite{wei2019sparse} and Sparse Flickering. For the works shown in Table \ref{tabel:related}, only \cite{wei2019sparse} is the white-box sparse attack; \cite{wei2020heuristic}\cite{yan2020sparse} are black-box sparse attack methods. As our work is a white-box sparse attack, we choose the most related one, Sparse \cite{wei2019sparse}, as the main baseline. We perform perturbation directly on the frame, while \cite{chen2019appending} appended additional frame in the end of video, which is more visible to human. So we did not include it as a baseline due to its compromise on the similarity of human perception. In \cite{Li_2019}, GANs are used to attack real-time video, which is not comparable to our method. We modified Flickering \cite{naeh2020flickering}, which perturbs all frames, into a sparse one as the Sparse Flickering baseline, but we still show the performance of perturbing all frames.

\textbf{Experiments Setting:}
The length of all input videos is crafted to be the same (40 frames). We randomly select 200 videos from different categories in the test dataset. For those experiments without saying the specific constraint, the maximum allowed search iteration (100 iterations) is applied; all experiments use Adam optimiser with 0.01 learning rate. The parameter $\lambda$ is set to 1.5 for the CNN+LSTM model, and 1.0 for the I3D and Inception-v3 models. For $\lambda$, values that can balance the fooling rate and perturbation strength are used.

\textbf{Metrics:}
 \textit{Fooling Rate (FR)}: the percentage of generated adversarial videos that are misclassified successfully. \textit{Average Number of Iterations (ANI)}: the average number of iterations taken to generate adversarial examples successfully based on the same original videos, which is used to measure the efficiency when we set a constraint on the maximum allowed iteration.

\begin{table*}\footnotesize
\begin{center}
\begin{tabular}{|c|c|c|c|c|c|}
\hline
\multirow{2}{*}{Models}&{\multirow{2}{*}{Attack Method}}&\multicolumn{2}{c|}{UCF101}&\multicolumn{2}{c|}{HMDB51}\\
    \cline{3-6}
    {}&{}&FR&ANI&FR&ANI\\
\hline\hline
\multirow{4}*{CNN+LSTM} & Sparse & $\SI{52.77 \pm 2.44}{\percent}$ & 16.45  & $\SI{95.2 \pm 1.8}{\percent}$ & 16.4\\
 \cline{2-6}
{}&Sparse Flickering&$\SI{48.48 \pm 1.67}{\percent} $& 23.55  & $\SI{91.94 \pm 2.93}{\percent} $ &8.4  \\
 \cline{2-6}
{}&DeepSAVA(without BO)& $\SI{56.22 \pm 1.65}{\percent}$&\cellcolor[gray]{.9} 8.32&$\SI{99.27 \pm 0.34}{\percent}$&8.42\\
 \cline{2-6}
{}&DeepSAVA(BO)& \cellcolor[gray]{.9}$\SI{57.22 \pm 1.36}{\percent}$&8.77&\cellcolor[gray]{.9}100\%&\cellcolor[gray]{.9}6.6\\
\hline\hline
\multirow{4}*{I3D} &Sparse & $\SI{10.12 \pm 1.19}{\percent}$&44&$\SI{5.74 \pm 1.25}{\percent}$&25.1\\
\cline{2-6}
{}&Sparse Flickering&$\SI{1.15 \pm 0.68}{\percent}$&13 &0\% &- \\
 \cline{2-6}
{}&DeepSAVA(without BO)& $\SI{47.57 \pm 2.64}{\percent}$&12.15&$\SI{46.39 \pm 3.86}{\percent}$&12.2\\
 \cline{2-6}
{}&DeepSAVA(BO)& $\cellcolor[gray]{.9}\SI{99.89 \pm 0.11}{\percent}$&\cellcolor[gray]{.9}6.47&$\cellcolor[gray]{.9}\SI{99.92 \pm 0.08}{\percent}$&\cellcolor[gray]{.9}5.35\\
\hline\hline
\multirow{4}*{Inception-v3} & Sparse & $\SI{42.25 \pm 4.30}{\percent}$&33.70&$\SI{45.82 \pm 1.56}{\percent}$&22.06\\
\cline{2-6}
{}&Sparse Flickering& $\SI{21.73 \pm 1.39}{\percent}$&35.4 &$\SI{27.55 \pm 0.98}{\percent}$ & 27.25 \\
 \cline{2-6}
{}&DeepSAVA(without BO)& $\SI{68.86 \pm 1.83}{\percent}$&13.29&$\SI{68.98 \pm 3.19}{\percent}$&11.84\\
 \cline{2-6}
{}&DeepSAVA(BO)& $\cellcolor[gray]{.9}\SI{70.39 \pm 2.78}{\percent}$&\cellcolor[gray]{.9}10.52&$\cellcolor[gray]{.9}\SI{74.74 \pm 0.82}{\percent}$&\cellcolor[gray]{.9}9.07\\
\hline
\end{tabular}
\end{center}
\caption{Comparison with baselines, DeepSAVA without BO and with BO on different models by only perturbing one frame. '-' means that there is no successful attack. Gray cell shows the best results. }

\label{tabel:result}
\end{table*}

\subsection{Comparison With Baseline Methods} 
 In this section, we will show the comparison results between DeepSAVA and baselines. Since running BO will add extra time to choose the frame, to make the comparison more complete, we also take the DeepSAVA without BO selection into account. 

\textbf{Limited iterations:}
Since each method uses a different metric, in order to control the maximum allowed perturbation we limit the number of search iterations for all methods. Each iteration only allows a small amount of perturbation (controlled by the learning rate of Adam optimiser), following the same setup used by the baselines. The results in Table \ref{tabel:result} show that the ANIs are much below the maximum allowed iteration (100), and we also found that even when it reaches the maximum iteration, the $lp$-norm and $SSIM$ distances are still acceptable \ref{label:maxiter}. Given that, setting a constraint on the maximum search number to 100 will not lead to large distortion. \begin{table}[H]\footnotesize
\begin{center}
\begin{tabular}{|c|c|c|c|c|c|}
\hline
max iter & FR & max(lp)&max(ssim)&ave(lp)&ave(ssim)\\
\hline
 30 & 0.5 & 0.11& 0.094 & 0.052 & 0.069  \\
\hline
50 & 0.5 & 0.135& 0.094 & 0.059 & 0.099   \\
\hline
80 &0.529 & 0.131 & 0.0959 & 0.0595 & 0.081\\
\hline
100 &0.529 & 0.131 & 0.095 & 0.052 & 0.067\\
\hline
\end{tabular}
\end{center}
\caption{\small{The relationship between the iteration, $l_{1,2}$, SSIM, and Fooling Rate for the I3D model with combine perturbation on UCF101.}}
\label{label:maxiter}
\vspace{-8mm}
\end{table}

We run the experiments 10 times, and show the average results with a 99\% confidence interval. For the methods without frame selection, the first frame is perturbed.As shown in Table \ref{tabel:result}, BO selection is more efficient than the one without BO. This happens because it is able to select the most critical frame, which can improve the efficiency on most of the cases. For the CNN+LSTM model, DeepSAVA increases the FR slightly compared with the baselines; while for the I3D model, we can see that the FR grows significantly. The BO selection process is also essential for I3D. Without BO, only about half of the test videos can be attacked successfully; after applying BO, the FR increases to nearly 100\%. As for the Inception-v3 model, the FR increases when applying DeepSAVA. It can be concluded that the CNN+LSTM is the most robust classification model among the three classifiers. Although the I3D has the highest classification accuracy, it is more vulnerable to attacks, even when only one frame is modified. That might happen because the I3D model relies heavily on the integral structure of the video itself and some frames may be more important.

We find that the position of key frames is related to the classifiers evaluated: for CNN+LSTM, the frames in the front are more often selected, and for other CNN networks, the position is variant. Thus, it is reasonable that the BO cannot improve the FR for CNN+LSTM model as much as the I3D, as we attacked the first frame when not selecting. We also show the results in Figure \ref{Fig:frame} for attacking a different number of frames across I3D and CNN+LSTM models. It can be seen that the more frames attacked, the higher the fooling rate obtained.
 \begin{figure}
\centering
\subfloat[CNN+LSTM]{
\begin{minipage}{6cm}
\centering
\includegraphics[width=6cm]{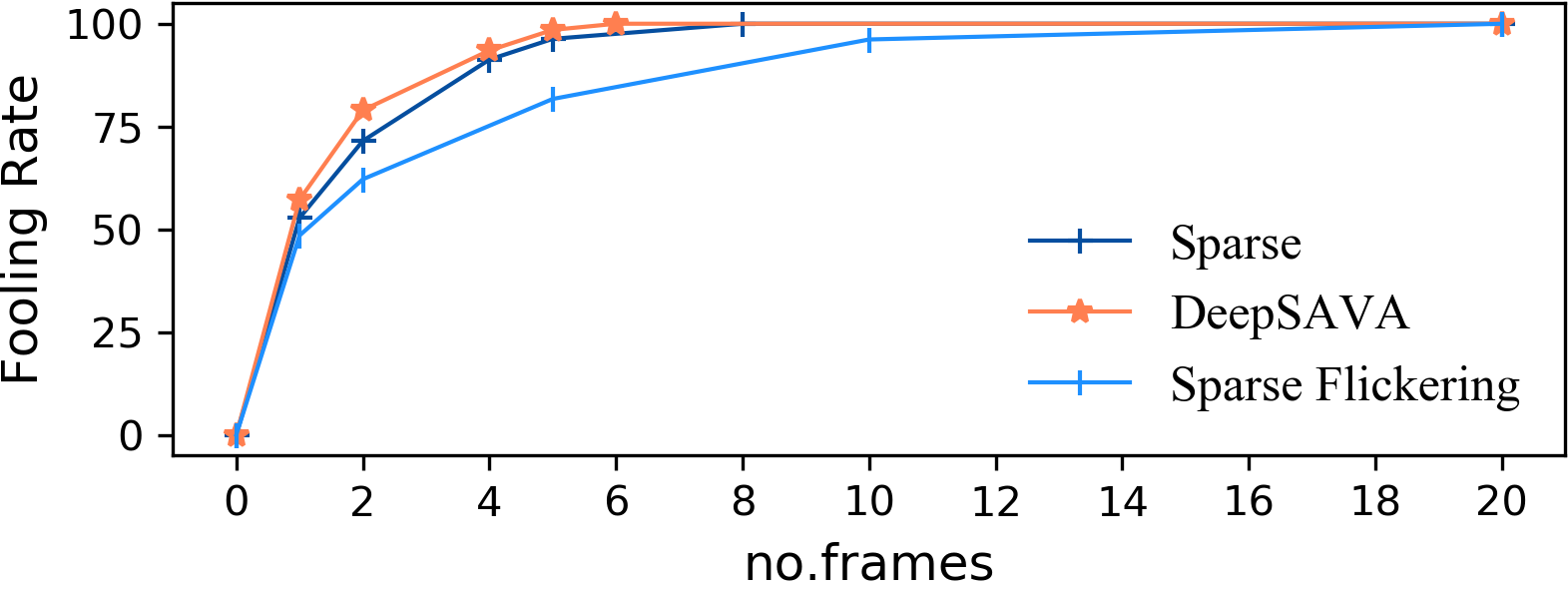}
\end{minipage}%
}%
\subfloat[I3D]{
\begin{minipage}{6cm}
\centering
\includegraphics[width=6cm]{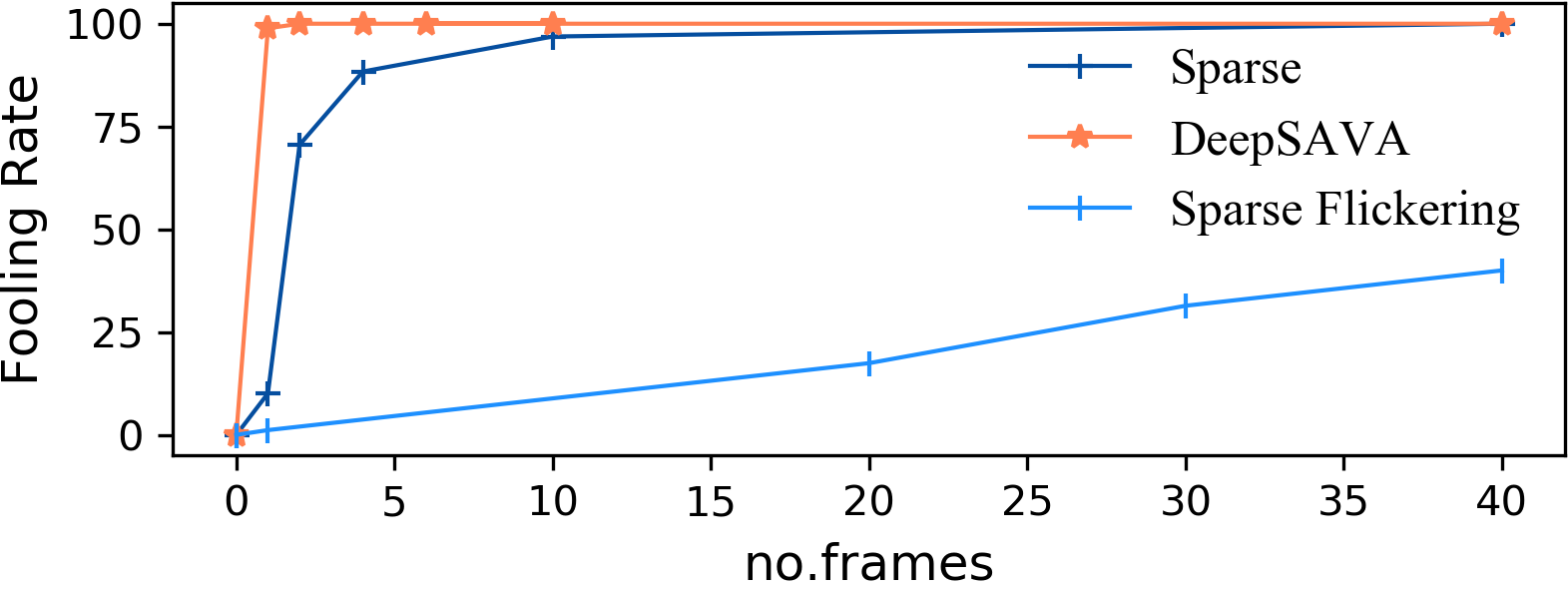}
\end{minipage}%
}
\subfloat[Inception-v3]{
\begin{minipage}{6cm}
\centering
\includegraphics[width=6cm]{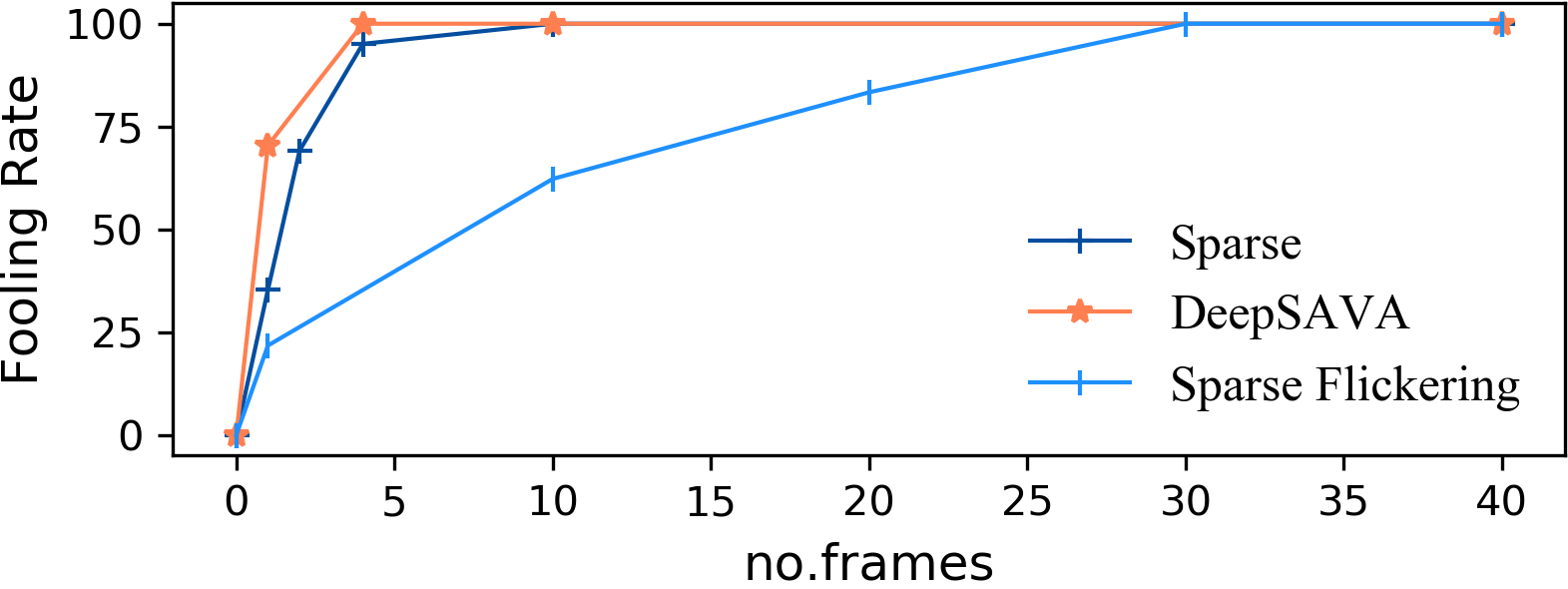}
\end{minipage}%
}%
\caption{Fooling Rate of attacking different number of frames across different models.}

\label{Fig:frame}
\end{figure}
 
\textbf{Fixed $l_{2,1}$ norm and SSIM:} For the purpose of a fair comparison, we also present the results under fixed $l_{2,1}$ and SSIM budgets for perturbing only one frame. The maximum allowed iteration is set to be 500 to limit the time. As the baseline methods are based on $l_p$-norm and our method is on SSIM, we take experiments under the same $l_p$-norm constraint and SSIM constraint, respectively. Based on the results of fixed iterations, we randomly select 200 videos from different categories to attack the I3D model on the UCF101 dataset. During the experiments, the Sparse Flickering spent days to achieve the constraint, thus we will only compare with the Sparse \cite{wei2019sparse} attack. In \cite{8743213}, the SSIM budget for attacking image is set to 0.95, thus we choose the SSIM constraints above 0.95. In \cite{5466111}, it states that the difference between the images is imperceptible when the $l_{2,1}$ score is 4, given that, we also set the $l_{2,1}$-norm budget to below $0.1$ (since $0.1*40 = 4$, as we have $40$ frames). As we can see in Table \ref{label:fixed budget}, under small fixed budgets, DeepSAVA outperforms the Sparse \cite{wei2019sparse} in both cases in terms of FR and total time.

 \begin{table}\footnotesize
\begin{center}
 \scalebox{0.9}{
\begin{tabular}{|c|c|c|c|c|c|c|}
\hline
  \multicolumn{7}{|c|}{$l_{2,1}$-norm}\\
\hline
Constraint & \multicolumn{3}{c|}{$l_{2,1}$ budget = 0.08} & \multicolumn{3}{c|}{$l_{2,1}$ budget = 0.09}\\
\hline
 Method & Sparse & DeepSAVA(no BO)&DeepSAVA & Sparse & DeepSAVA(no BO) & DeepSAVA  \\
\hline
FR & 40.51\% & 48.1\% &\cellcolor[gray]{.9} 88.61\% & 41.77\% & 54.43\%&  \cellcolor[gray]{.9}93.67\% \\
\hline
Time (s) & 8018.9 & 2629 & \cellcolor[gray]{.9}1535.8 & 14001 & 3729 & \cellcolor[gray]{.9} 1573.82\\
\hline
  \multicolumn{7}{|c|}{$SSIM$}\\
\hline
Constraint & \multicolumn{3}{c|}{$SSIM$ budget = 0.98} & \multicolumn{3}{c|}{$SSIM$ budget = 0.96}\\
\hline
 Method & Sparse & DeepSAVA(no BO)&DeepSAVA & Sparse & DeepSAVA(no BO) & DeepSAVA  \\
\hline
FR & 8.06\% & 16.56\% &\cellcolor[gray]{.9} 35.44\% & 10.1\% & 51.9\%& \cellcolor[gray]{.9} 96.20\%  \\
\hline
Time (s) & 5842.32 & \cellcolor[gray]{.9}1285.1 & 1424.4 & 13789.23 & 5633.28 & \cellcolor[gray]{.9}1545.5\\
\hline
\end{tabular}
}
\end{center}
\caption{Attack I3D model on UCF101 dataset under $l_{2,1}$ and $SSIM$ constraint separately.}
\label{label:fixed budget}
\end{table}

\begin{figure}\small
\centering
\subfloat[]%
[\small{apply eye makeup}]%
{\includegraphics[width=2.5cm]{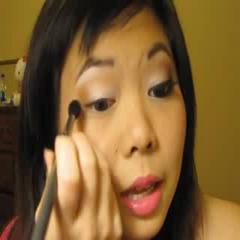}}
\subfloat[]%
[\textcolor{red}{apply-lips}\\(DeepSAVA)]%
{\includegraphics[width=2.5cm]{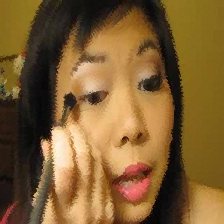}}
\subfloat[]%
[\textcolor{red}{apply-lips}\\(Sparse)]%
{\includegraphics[width=2.5cm]{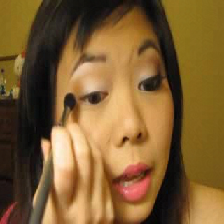}}
\subfloat[]%
[ParallelBars]%
{\includegraphics[width=2.5 cm]{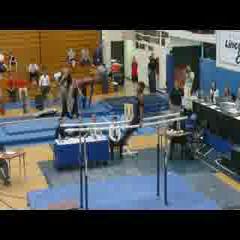}}
\subfloat[]%
[\textcolor{red}{Haircut}\\ (DeepSAVA)]%
{\includegraphics[width=2.5cm]{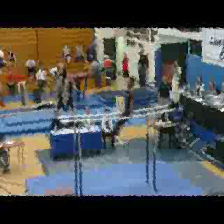}}
\subfloat[]%
[\small{\textcolor{red}{Haircut} (Sparse)}]%
{\includegraphics[width=2.5cm]{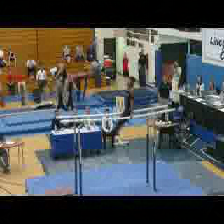}}
\caption{Original, and adversarial examples generated by DeepSAVA and Sparse \cite{wei2019sparse} when only one frame in the video is perturbed. The red labels are the wrong predictions.}
\label{Fig:Results}
\end{figure}
\subsection{ {Average Absolute Perturbation}}
 Average Absolute Perturbation (AAP) is introduced to measure the perturbation level for each method. As mentioned previously, the sparse Flickering adds a small perturbation per frame, but cannot obtain comparable results to ours. Thus, we choose the pure sparse attack (Sparse) as the main baseline to show the average absolute perturbation. As the baseline is guided by $l_{1,2}$ norm and ours is based on SSIM loss, we will record the average perturbation of $l_{1,2}$ and SSIM separately. To achieve a fair comparison, we set the maximum $l_{1,2}$ norm ball constraint as 0.1 and maximum SSIM constraint as 0.92. Suppose the fooling rate is $f$, and distant matrix is $D$, which can be set to (1-SSIM) or $l_{1,2}$ norm, thus the average absolute perturbation(AAP) can be represented as:
  $$AAP(D)=\frac{\sum_N D(V_{adv}-V_{original})}{N}*f+ D_{max}*(1-f),$$
  where $V_{adv}$ denotes the generated adversarial video that could successfully mislead the classifier and $D_{max}$ is the maximum constraint; $N$ is the number of adversarial samples achieving successful attack. We run experiments on 200 random selected videos of UCF101 dataset and record the results of FR, ANI, AAP($l_{1,2}$) and AAP(SSIM) in Table \ref{tabel:aap}. From the results, we can see that for the model I3D and Inception-v3, our method could achieve better performance in terms of efficiency, fooling rate and AAP(SSIM). For the CNN+LSTM model, our method engages higher fooling rate and spends less time, and the AAP($l_{1,2}$) and AAP(SSIM) are also acceptable compared with the baseline model.
 \begin{table*}\footnotesize
\begin{center}
\begin{tabular}{|c|c|c|c|c|c|}
\hline
\multirow{2}{*}{Models}&{\multirow{2}{*}{Attack Method}}&\multicolumn{4}{c|}{UCF101}\\
    \cline{3-6}
    {}&{}&FR&ANI&AAP($l_{1,2}$)&AAP(SSIM)\\
\hline\hline
\multirow{3}*{CNN+LSTM} & Sparse & {54.31\%} & {15.31} & {\cellcolor[gray]{.9}0.054} & {\cellcolor[gray]{.9}0.043} \\
 \cline{2-6}
 \cline{2-6}
{}&DeepSAVA(without BO)& {56.94\%}&{\cellcolor[gray]{.9}7.87} & {0.077} & {0.060}\\
 \cline{2-6}
{}&DeepSAVA(BO)& {\cellcolor[gray]{.9}57.11\%} & {8.01} & {0.071} & {0.058}\\
\hline\hline
\multirow{3}*{I3D} &Sparse & {11.22\% }&{49} & {0.092 }& {0.079}\\
\cline{2-6}
{}&DeepSAVA(without BO)&{ 48.78\%}  & {11.34} &{ 0.0857} & {0.054}  \\
 \cline{2-6}
{}&DeepSAVA(BO)&{ \cellcolor[gray]{.9}99.89\%} &{\cellcolor[gray]{.9}5.74} & {\cellcolor[gray]{.9}0.055} & {\cellcolor[gray]{.9} 0.0233} \\
\hline\hline
\multirow{3}*{Inception-v3} & Sparse & {41.84\%} &{38.21}&{\cellcolor[gray]{0.9}0.062} & {0.0512}\\
\cline{2-6}
{}&DeepSAVA(without BO)& {65.14\%}&{14.88} & {0.072 }& {0.052}\\
 \cline{2-6}
{}&DeepSAVA(BO)& {\cellcolor[gray]{.9}77.49\% }& {\cellcolor[gray]{.9}11.43} & {0.071 }& {\cellcolor[gray]{0.9}0.0508}\\
\hline
\end{tabular}
\end{center}
\caption{{Comparison with Sparse baseline, DeepSAVA without BO and with BO on different models by only perturbing one frame. Gray cell shows the best results. }}
 \vspace{-3mm}
\label{tabel:aap}
\end{table*}
\subsection{Visualization of Results}
The generated adversarial frames by DeepSAVA are presented in Figure \ref{Fig:Results}. Because of the spatial transformation, the frame looks a little bit shaky but not obvious in human eyes. In fact, in the real world, it is normal to see that there are a few frames with instabilities during video shooting and transmitting. That's why we apply the spatial transformation in video attacks to improve the efficiency and fooling rate. In practice, a distortion in one frame of a video is less noticeable than a static image since this specific frame only appears for 0.047 seconds in human eyes \cite{xiao2018generating}. We could also see that it does not lead to a noticeable perturbation as shown by our video demos.

When transmitting the videos in the real world, the generated frames need to be compressed into videos first and then decompressed to frames. We found that for the additive-only perturbed frames, they may not remain adversarial examples after such transmission. Our experiments demonstrate that DeepSAVA can be immune to short video compression due to the fact that perturbation based on spatial transformation can be well preserved during compression while additive perturbation may disappear. 

\subsection{Ablation Study}
\begin{table*}[htb]\footnotesize
\begin{center}
\begin{tabular}{|l| c c |c|c|}
\hline
\multirow{2}{*}{Approach}&\multicolumn{2}{c|}{CNN+LSTM}&\multicolumn{1}{c|}{Inception-v3}&\multicolumn{1}{c|}{I3D}\\
    \cline{2-5}
    {}&Mask1&Mask4&Mask1&Mask1\\
\hline\hline
Fixed the frame (first $n$-th)&{}&{}&{}&{}\\
$\mathcal{D}$ &$\SI{52.77 \pm 2.24}{\percent}$ & $\SI{91.28 \pm 1.95}{\percent}$&$\SI{42.45 \pm 4.30}{\percent}$&$\SI{10.12 \pm 1.19}{\percent}$ \\
$\mathcal{S}$ & $\SI{55.27 \pm 1.82}{\percent}$ & $\SI{91.89 \pm 1.45}{\percent}$ &$\SI{63.91 \pm 5.61}{\percent}$& $\SI{29.99 \pm 2.36}{\percent}$\\
$\mathcal{D} + \mathcal{S}$ &$\SI{56.22 \pm 1.65}{\percent}$ &$\SI{92.99 \pm 1.85}{\percent}$&$\SI{68.86 \pm 1.83}{\percent}$ & $\SI{47.57 \pm 2.64}{\percent}$\\
\hline
Using BO to choose frame&{}&{}&{}&{}\\
$\mathcal{D} + \mathcal{S}$ &$\cellcolor[gray]{.9}\SI{57.22 \pm 1.36}{\percent}$ &\cellcolor[gray]{.9}$\SI{93.51 \pm 1.33}{\percent}$ &\cellcolor[gray]{.9}$\SI{70.39 \pm 2.78}{\percent}$& \cellcolor[gray]{.9}$\SI{99.89 \pm 0.11}{\percent}$\\
\hline
\end{tabular}
\end{center}
\caption{Effects of combining noise ($\mathcal{D}$) and spatial transformation ($\mathcal{S}$) by modifying a different number of frames on UCF101; Mask $N$ means that $N$ frames are modified.}
\label{table:combine}
\end{table*}
We perform ablation experiments to study the effects of combined perturbation for a different number of attacked frames by comparing with additive noise only and spatial transform only perturbations, and the effects of BO selection. Table \ref{table:combine} shows the FR for three classifiers on the UCF101 dataset. Four approaches are taken to attack the model: 1) only noise ($\mathcal{D}$), 2) only spatial transformation ($\mathcal{S}$), 3) combination of additive perturbation and spatial transformation ($\mathcal{D} + \mathcal{S}$), and 4) combined perturbation with BO selection. To make more comprehensive evaluations on the superiority of combination, we attack a different number of frames for the CNN+LSTM model as it has the lowest FR when only perturbing one frame. All experiments showed the combination power to increase the FR; using BO selection is also useful, especially for the I3D model. 
\subsection{{The Accuracy of Bayesian Optimisation Selection}}
To justify whether the Bayesian Optimisation could select the most critical frames, we take the brute force search experiments to obtain the upper bound of the performance: when the selection frame is 1, we select the key frame manually one by one of the video, and then record the maximum loss found by the search. We randomly select 100 videos from UCF101 in different categories. The fooling rates, average maximum loss and average time spent for one video on three models are shown in the Table \ref{tabel:upper bound}. From the results, we can obtain the same results as the brute force search, but spending much less time, which confirms the effectiveness of BO optimisation.
 \begin{table*}[htb]\footnotesize
\begin{center}
\begin{tabular}{|c|c|c|c|c|c|c|c|}
\hline
\multirow{2}{*}{Approach} &\multicolumn{2}{c|}{CNN+LSTM}& \multicolumn{2}{c|}{Inception-v3}& \multicolumn{2}{c|}{I3D}&\multirow{2}{*}{time (s)}\\
\cline{2-7}
{}&FR&average loss &FR &average loss &FR &average loss&\\
\hline
BO Selection&{55.81\%} &{0.21}&{72.22\%}&{3.39}& {100\%}&{1.35}&{16.1}\\
\hline
brute force search &{55.81\%}&{0.27 }&{72.22\%} &{3.39}&{100\%}&{1.35}&{70.4}\\
\hline
\end{tabular}
\end{center}
\caption{{Fooling Rate, average selected maximum loss and average time spent for one video of BO Selection and brute force search.}}
 \vspace{-3mm}
\label{tabel:upper bound}
\end{table*}
\subsection{Transferability across recurrent models}
\begin{table}[htb]\footnotesize
\begin{center}
\scalebox{0.9}{
\begin{tabular}{|c| c| c |c |c |c| c| c| c| c| c| c|}
\hline
\multirow{2}{*}{Models}  & \multicolumn{2}{c|}{LSTM} & \multicolumn{2}{c|}{Vanilla RNN} &
\multicolumn{2}{c|}{GRU} & \multicolumn{2}{c|}{Inception-v3} & \multicolumn{2}{c|}{I3D}\\
\cline{2-11}
 & Sparse & DeepSAVA &Sparse & DeepSAVA&Sparse & DeepSAVA &Sparse & DeepSAVA &Sparse & DeepSAVA \\
\hline
LSTM& 100\% & 100\% &34.42\% & \cellcolor[gray]{.9}41.38\% & 64.35\%&\cellcolor[gray]{.9}85.34\% & 50.0\%&\cellcolor[gray]{.9}52.17\% & 53.48\%&\cellcolor[gray]{.9}54.62\%\\
\hline
{Vanilla RNN} & \cellcolor[gray]{.9}100\% &\cellcolor[gray]{.9}100\% &\cellcolor[gray]{.9}100\%&\cellcolor[gray]{.9}100\%&\cellcolor[gray]{.9}100\% &\cellcolor[gray]{.9}100\% & 71.74\%&\cellcolor[gray]{.9}82.40\% & 60.50\%&\cellcolor[gray]{.9}64.02\%  \\
\hline
{GRU} &79.34 \% & \cellcolor[gray]{.9}84.75\%& 40.70\%& \cellcolor[gray]{.9}56.03\% &100\% & 100\% & 50.0\%&\cellcolor[gray]{.9}51.08\%& 42.68\%&\cellcolor[gray]{.9}49.58\%\\
\hline
{Inception-v3} &22.95 \% & \cellcolor[gray]{.9}24.36\%& 22.80\%& \cellcolor[gray]{.9}26.72\% &22.90\% & \cellcolor[gray]{.9}31.03\% & \cellcolor[gray]{.9}100\%&\cellcolor[gray]{.9}100\%& 33.61\%&\cellcolor[gray]{.9}37.80\%\\
\hline
{I3D} &6.56 \% & \cellcolor[gray]{.9}10.08\%& 7.01\%& \cellcolor[gray]{.9}9.48\% &7.64\% & \cellcolor[gray]{.9}8.62\% & 13.04\%&\cellcolor[gray]{.9}14.13\%& \cellcolor[gray]{.9}100\%&\cellcolor[gray]{.9}100\%\\
\hline
\end{tabular}}
\end{center}
\caption{\small{Fooling Rate across recurrent models on UCF101.}}
\label{label:Transferability}
\end{table}
The transferability across models is an important evaluation of adversarial attacks, which can be treated as a black-box problem without accessing parameters of the target model.
In our work, the I3D and Inception-v3 only contain CNN, while the recurrent neural networks (RNN) like CNN+LSTM contains the time-related network. Due to the unique time-related structure of videos, we mainly focus the transferability across time-related networks. We perform the transferability experiments on the UCF101 dataset for different RNNs, Inception-v3 and I3D models. The features of original videos are extracted firstly by CNN (Inception-v3) model and then are fed into vanilla RNN \cite{rumelhart1986learning}, LSTM \cite{lstm}, and GRU \cite{cho-etal-2014-learning} networks respectively. The training accuracy for vanilla RNN and GRU are 65.16\% and 73.05\% respectively.

As Figure \ref{Fig:frame} shows that the Sparse \cite{wei2019sparse} performs better than the Sparse Flickering in terms of FR, we choose the Sparse \cite{wei2019sparse} as the baseline method. The fooling rates (FR) of the generated videos across models are presented in Table \ref{label:Transferability}. The models in rows are used to generate adversarial videos and the models in columns are the target attack classifiers. Here we disturb seven frames of a video to enlarge the attacking success rate. We use the adversarial examples generated from the white-box attack for the transferability, which leads to the FR in the diagonal being 100\%. These adversarial examples are then used to attack other models (like a black-box attack) as detailed in Table \ref{label:Transferability}. Comparing with the baseline, our approach has a higher FR which indicates a better performance in terms of transferability. The difference between vanilla RNN and the other models is that vanilla RNN has no memory component, so it shows a weak performance on the video classification task. As we observed, adversarial videos generated from LSTM and GRU models can fool the vanilla RNN successfully. Additionally, the FR across GRU and LSTM are around 85\%, which shows good transferability between the recurrent models with memory. However, from the fooling rates shown in Table \ref{label:Transferability}, we could see that the transferability from RNNs to CNNs is not as good as that from CNNs to RNNs. the fooling rates to attack RNN models are higher than those to attack the I3D and Inception-v3 models. While that happens maybe because the I3D model has the highest training accuracy, it engages the lowest fooling rate when performing the black-box attack on it. As we can also conclude that due to the lowest training accuracy the Vanilla RNN model obtains, it achieves the highest fooling rate on attacking unseen Vanilla RNN model. Overall, compared with the Sparse baseline, our method could achieve better transferability.  

\section{Conclusion}

In this paper, we apply spatial transformed perturbation and additive noise to attack as few frames as possible to obtain the sparse adversarial videos. We run experiments on the UCF101 and HMDB51 action dataset and three models. The most influential frames to be attacked are selected by a joint optimisation strategy with Bayesian optimisation (BO) and SGD-based optimisation.
Additionally, the quality of generated adversarial examples is measured by SSIM instead of $l_p$-norm. We obtain better results than state-of-the-art sparse baselines in terms of both fooling rate and transferability. Our most significant results are for the I3D model, by only attacking one frame of the video to obtain 99.5\% to 100\% attack success rate.

\textbf{ Acknowledgements}
This work was supported by Partnership Resource Fund (PRF) \textit{Towards the Accountable and Explainable Learning-enabled Autonomous Robotic Systems (AELARS)}, funded via the UK EPSRC projects on Offshore Robotics for Certification of Assets (ORCA) [EP/R026173/1].
This work was also funded by the Faculty of Science and Technology of Lancaster University. We also thank the High End Computing facility at Lancaster University for the computing resources, and Abdulrahman Kerim and Washington L. S. Ramos for the thorough reviews.
{\small
\bibliographystyle{ieee_fullname}
\bibliography{bibvideo.bib}
}
\begin{appendix}

\section{Calculating the Gradient of SSIM}

The SSIM was first proposed in \cite{wang2002universal}, and is detailed in \cite{wang2004image}. Given $x$ and $\hat{x}$ as the local pixels taken from the same location of the same frame in the clean video and adversarial video, respectively, the local similarity between them can be computed on three aspects: structures ($s(x,\hat{x})$), contrasts ($c(x,\hat{x})$), and brightness values ($b(x,\hat{x})$). The local SSIM is formed by these terms \cite{wang2004image}:
\begin{equation}
\begin{aligned} &S(x, \hat{x})= s(x, \hat{x}) \cdot c(x, \hat{x}) \cdot b(x, \hat{x})=
\\ &\left(\frac{\sigma_{x \hat{x}}+D_{1}}{\sigma_{x} \sigma_{\hat{x}}+D_{1}}\right)  \cdot\left(\frac{2 \sigma_{x} \sigma_{\hat{x}}+D_{2}}{\sigma_{x}^{2}+\sigma_{\hat{x}}^{2}+D_{2}}\right) \cdot \left(\frac{2 \mu_{x} \mu_{\hat{x}}+D_{3}}{\mu_{x}^{2}+\mu_{\hat{x}}^{2}+D_{3}}\right),
\end{aligned}
\end{equation}

The structural similarity index (SSIM) measure in Equation (2) can be expressed as: \cite{wang2004image}
\begin{equation}
    \operatorname{SSIM}(\mathbf{x}, \mathbf{\hat{x}})=\frac{\left(2 \mu_{x} \mu_{\hat{x}}+C_{1}\right)\left(2 \sigma_{x \hat{x}}+C_{2}\right)}{\left(\mu_{x}^{2}+\mu_{\hat{x}}^{2}+C_{1}\right)\left(\sigma_{x}^{2}+\sigma_{\hat{x}}^{2}+C_{2}\right)}
\end{equation}
The mean of $x$ , the variance of $x$, and co-variance of $x$ and $\hat{x}$ can be represented as $\mu_{x}$, $\sigma_{x}^{2}$ and $\sigma_{x \hat{x}}$. They can be calculated respectively:
\begin{equation}
\begin{aligned}
&\mu_{x}= \frac{1}{N_{P}}\left(\mathbf{1}^{T} \cdot \mathbf{x}\right)\\&\sigma_{x}^{2}=\frac{1}{N_{P}-1}\left(\mathbf{x}-\mu_{x}\right)^{T}\left(\mathbf{x}-\mu_{x}\right)\\&\sigma_{x \hat{x}} =\frac{1}{N_{P}-1}\left(\mathbf{x}-\mu_{x}\right)^{T}\left(\mathbf{\hat{x}}-\mu_{\hat{x}}\right) 
\end{aligned}
\end{equation}
Given $x$ and $\hat{x}$ as the local pixels taken from the same location of the same frame in the clean video and adversarial video, respectively, the local similarity between them can be computed on three aspects: structures ($s(x,\hat{x})$), contrasts ($c(x,\hat{x})$), and brightness values ($b(x,\hat{x})$).
The local SSIM is formed as \cite{wang2004image}:
\begin{equation}\begin{aligned} &S(x, \hat{x})= s(x, \hat{x}) \cdot c(x, \hat{x}) \cdot b(x, \hat{x})=
\\ &\left(\frac{\sigma_{x \hat{x}}+D_{1}}{\sigma_{x} \sigma_{\hat{x}}+D_{1}}\right)  \cdot\left(\frac{2 \sigma_{x} \sigma_{\hat{x}}+D_{2}}{\sigma_{x}^{2}+\sigma_{\hat{x}}^{2}+D_{2}}\right) \cdot \left(\frac{2 \mu_{x} \mu_{\hat{x}}+D_{3}}{\mu_{x}^{2}+\mu_{\hat{x}}^{2}+D_{3}}\right),\end{aligned}
\end{equation}
where $\mu_{x}$ and $\mu_{\hat{x}}$ denote means, $\sigma_{x}$ and $\sigma_{\hat{x}}$ are standard deviations of $x$ and $\hat{x}$, respectively; $\sigma_{x\hat{x}}$ represents the cross correlation of $x$ and $\hat{x}$ after deleting means; D1, D2, and D3 are weight parameters. For SSIM metric, a value of 1 means that the two images compared are the same. 
As the SSIM is calculated based on pixel level, it use a sliding window method, which moves pixel by pixel by the window across the whole image. As we use uniform pooling to combine the total SSIM for the whole videos, suppose we have $N$ pixels in the total videos, the SSIM can be represented as:
\begin{equation}
\operatorname{SSIM}(\mathbf{X}, \mathbf{\hat{X}})=\frac{\sum_{i=1}^{N}  \cdot \operatorname{SSIM}\left(\mathbf{x}_{i}, \mathbf{\hat{x}}_{i}\right)}{N}
\end{equation}
where $x_i$ and $\hat{x}_i$ are the $i$-th pixel of each frame in the video.
To apply the gradient decent optimisation method described in Section 3, we have to compute the gradient of SSIM with respect to the adversarial video example $\mathbf{\hat{X}}$. As equation (9) shows, to compute $\vec{\nabla}_{\mathbf{\hat{X}}} \operatorname{SSIM}(\mathbf{X}, \mathbf{\hat{X}})$, we only need to calculate the gradient $\vec{\nabla}_{\hat{x}_i} \operatorname{SSIM}(x_i, \hat{x}_i)$. The process is represented as follows. \cite{wang2004stimulus}
We define four parameters to deduce the derivative of local SSIM:
\begin{equation}
\begin{array}{l}M_{1}=2 \mu_{x} \mu_{\hat{x}}+C_{1}, \quad M_{2}=2 \sigma_{x \hat{x}}+C_{2} \\ P_{1}=\mu_{x}^{2}+\mu_{\hat{x}}^{2}+C_{1}, \quad P_{2}=\sigma_{x}^{2}+\sigma_{\hat{x}}^{2}+C_{2}\end{array}
\end{equation}
Therefore, the gradient can be expressed as:
\begin{equation}
\vec{\nabla}_{\hat{x}} \operatorname{SSIM}(x, \hat{x})=\frac{2}{N_{P} P_{1}^{2} P_{2}^{2}}\left[M_{1} P_{1}\left(M_{2} x-P_{2} \hat{x}\right)+P_{1} P_{2}\left(M_{2}-M_{1}\right) \mu_{x}+M_{1} M_{2}\left(P_{1}-P_{2}\right) \mu_{\hat{x}}\right]
\end{equation}
\section{{Effects of $\lambda$} }
 {To decide the value of $\lambda$, we applied the DeepSAVA without BO selection on 200 random selected videos of UCF101 dataset to evaluate the effect of $\lambda$. The average success perturbation (ASP) is the average of the SSIM score of perturbation for the adversarial examples that could mislead the model successfully}:
{ $$ASP(SSIM))=avg(SSIM(V_{adv}-V_{original})),$$
 where $V_{adv}$ denotes the generated adversarial video that could successfully mislead the classifier and $V_{original}$ is the original video.}
 {The results of applying $\lambda = 0.8,1.0,1.5$ on three models are presented in Table \ref{tabel:lambda}. We can see that the bigger the $\lambda$, the lower the FR while the lower the perturbation.While, for the CNN+LSTM model, the fooling rate remains the same across all tested $\lambda$ values, but the perturbation level is the lowest at $\lambda$ = 1.5. Thus, we choose $\lambda =1.5$ for the CNN+LSTM model and $\lambda =1.0$ for I3D and Inception-v3 model to trade off the performance in terms of the fooling rate and average success perturbation.}
  \begin{table*}\footnotesize
\begin{center}
\begin{tabular}{|c|c|c|c|}
\hline
\multirow{1}{*}{Models}&{\multirow{1}{*}{$\lambda$ value}}&FR&ASP(SSIM)\\
\hline\hline
\multirow{3}*{CNN+LSTM} &0.8& 56.94\% &0.0429 \\
 \cline{2-4}
{}&1.0& 56.94\%&0.0412\\
 \cline{2-4}
{}&1.5&56.94\%&0.0401 \\
\hline\hline
\multirow{3}*{I3D} &0.8& 51.22\% &0.0316\\
\cline{2-4}
{}&1.0& 48.78\% & 0.0268\\
 \cline{2-4}
{}&1.5& {48.17\%}& {0.0198} \\
\hline\hline
\multirow{3}*{Inception-v3} &{0.8} & {66.05\%} & {0.0534}\\
\cline{2-4}
{}&{1.0}& {65.14\%}&{0.0518}\\
 \cline{2-4}
{}&{1.5}&{64.22\%} &{0.0454} \\
\hline
\end{tabular}
\end{center}
\caption{\small{{The results of DeepSAVA(without BO) on UCF101 dataset for different $\lambda$ values. }}}
 \vspace{-3mm}
\label{tabel:lambda}
\end{table*}
\section{Comparison Experiments with $l_p$-norm and SSIM Constraints}
 \subsection{I3D model}
\begin{table}[H]\footnotesize
\begin{center}
\begin{tabular}{|c|c|c|c|c|c|c|}
\hline
Constraint & \multicolumn{3}{c|}{$l_{2,1}$ budget = 0.1} & \multicolumn{3}{c|}{SSIM budget = 0.94}\\
\hline
 Methods & Sparse & DeepSAVA(no BO)&DeepSAVA & Sparse & DeepSAVA(no BO) & DeepSAVA  \\
\hline
FR & 60\% & 58.22\% &{\bf 91.25\% }& 12.9\% & 66.2\%& {\bf 97.46\%}   \\
\hline
Time (s) & 24029.8  & 4109.78 & {\bf 1483.96} & 30803.2 & 8276.1 & \bf{1586.3}\\
\hline
\end{tabular}
\end{center}
\caption{\small{Comparison with Sparse baseline method}}
\label{label:fixed budget}
\end{table}
 \subsection{CNN+LSTM model}
 \begin{table}[H]\footnotesize
\begin{center}
\begin{tabular}{|c|c|c|c|c|c|c|}
\hline
Constraint & \multicolumn{3}{c|}{$l_{2,1}$ budget = 0.08} & \multicolumn{3}{c|}{$l_{2,1}$ budget = 0.1}\\
\hline
 Methods & Sparse & DeepSAVA(no BO)&DeepSAVA & Sparse & DeepSAVA(no BO) & DeepSAVA  \\
\hline
FR & {\bf 55.71}\% & 48.57\% &50\% & {\bf 58.57}\% & {\bf 58.57}\%&  57.14\% \\
\hline
Time (s) & 22800.5 & {\bf 13777.6} & 15010 & 23336.8 & {\bf 19774.4} & 20866.4\\
\hline
\end{tabular}
\end{center}
\caption{Attack CNN+LSTM model on UCF101}
 \vspace{-3mm}
\label{label:fixed budget}
\end{table}
\begin{table}[H]\footnotesize
\begin{center}
\begin{tabular}{|c|c|c|c|c|c|c|}
\hline
Constraint & \multicolumn{3}{c|}{$SSIM$ budget = 0.96} & \multicolumn{3}{c|}{$SSIM$ budget = 0.94}\\
\hline
 Methods & Sparse & DeepSAVA(no BO)&DeepSAVA & Sparse & DeepSAVA(no BO) & DeepSAVA  \\
\hline
FR & {\bf 50}\% & 47.14\% & 47.14\% & {\bf 52.85}\% & {\bf 52.85}\%& {\bf 52.85\%}  \\
\hline
Time (s) & 15120.21 & {\bf 4039.24} & 5131.24 &  15952.52 & {\bf 6341.7} & 7433.3\\
\hline
\end{tabular}
\end{center}
\caption{Attack CNN+LSTM model on UCF101}
 \vspace{-3mm}
\label{label:fixed budget}
\end{table}
For the CNN+LSTM model, we can see that although the Sparse baseline could obtain higher fooling rate, but it will spent much more time to generate the adversarial examples. Our method using combined perturbation will spent less time and obtain comparable fooling rate. Because here we select the first frame to attack, the BO cannot improve the performance as I3D model. 
\end{appendix}
\end{document}